\documentclass{isprs} %
\usepackage{subfigure}
\usepackage{setspace}
\usepackage{geometry} %
\usepackage{epstopdf}
\usepackage[labelsep=period]{caption}  %
\usepackage{amsmath}
\usepackage[british]{babel} 
\usepackage[hang]{footmisc}
\usepackage{multirow}

\usepackage{booktabs}

\usepackage{graphicx}        %
\usepackage[final]{microtype} %
\usepackage{tikz}
\usepackage{xcolor}
\usepackage{stackengine}
\usepackage{algorithm}
\usepackage{algpseudocode}
\usepackage[hidelinks]{hyperref}
\usepackage{enumitem}
\usepackage{float}
\usepackage{stfloats}

\usepackage{adjustbox}

\usepackage{tabularx}
\newcolumntype{Y}{>{\centering\arraybackslash}X}

\usepackage{soul}

\algnewcommand{\LeftComment}[1]{{\color{gray}\Statex \(\triangleright\) #1}}

\begin{document}

\title{Diachronic Stereo Matching for Multi-Date Satellite Imagery}
\date{}

\author{Elías Masquil\textsuperscript{1}, 
Luca Savant Aira\textsuperscript{2},
Roger Marí\textsuperscript{3},
Thibaud Ehret\textsuperscript{4}, 
Pablo Musé\textsuperscript{1,5}, 
Gabriele Facciolo\textsuperscript{5,6}}

\address{
\textsuperscript{1 }IIE, Facultad de Ingeniería, Universidad de la República, Uruguay \\
\textsuperscript{2 }Politecnico di Torino, Corso Duca degli Abruzzi, Torino TO, Italia\\
\textsuperscript{3 }Eurecat, Centre Tecnologic de Catalunya, Multimedia Technologies, Barcelona, Spain \quad
\textsuperscript{4 }AMIAD, Pôle Recherche, France\\
\textsuperscript{5 }Universite Paris-Saclay, CNRS, ENS Paris-Saclay, Centre Borelli, Gif-sur-Yvette, France \quad 
\textsuperscript{6 }Institut Universitaire de France
}

\abstract{

Recent advances in image-based satellite 3D reconstruction have progressed along two complementary directions. On one hand, multi-date approaches using NeRF or Gaussian-splatting jointly model appearance and geometry across many acquisitions, achieving accurate reconstructions on opportunistic imagery with numerous observations. On the other hand, classical stereoscopic reconstruction pipelines deliver robust and scalable results for simultaneous or quasi-simultaneous image pairs.
However, when the two images are captured months apart, strong seasonal, illumination, and shadow changes violate standard stereoscopic assumptions, causing existing pipelines to fail.
This work presents the first Diachronic Stereo Matching method for satellite imagery, enabling reliable 3D reconstruction from temporally distant pairs. Two advances make this possible: (1) fine-tuning a state-of-the-art deep stereo network that leverages monocular depth priors, and (2) exposing it to a dataset specifically curated to include a diverse set of diachronic image pairs. In particular, we start from a pretrained MonSter model, trained initially on a mix of synthetic and real datasets such as SceneFlow and KITTI, and fine-tune it on a set of stereo pairs derived from the DFC2019 remote sensing challenge. This dataset contains both synchronic and diachronic pairs under diverse seasonal and illumination conditions. Experiments on multi-date WorldView-3 imagery demonstrate that our approach consistently surpasses classical pipelines and unadapted deep stereo models on both synchronic and diachronic settings.
Fine-tuning on temporally diverse images, together with monocular priors, proves essential for enabling 3D reconstruction from previously incompatible acquisition dates.
}

\keywords{Stereo Matching, 3D Reconstruction, Diachronic Matching, Multi-Date Satellite Images}

\maketitle

\vspace{1mm}

\begin{figure}[!ht]
    \centering
    {
    \small
    \newcommand{\maeOurs}{1.23} %
    \newcommand{\maeZero}{3.99} %
    \setlength{\tabcolsep}{0pt}
    \noindent\begin{tabularx}{\linewidth}{@{}YY@{}}
    \includegraphics[width=3cm, angle=180, trim=10mm 10mm 10mm 10mm, clip]{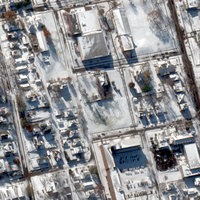} &
    \includegraphics[width=3cm, angle=180, trim=10mm 10mm 10mm 10mm, clip]{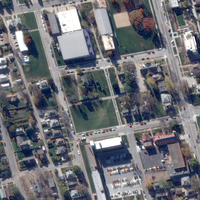} \\
    Left image (winter) &
    Right image (autumn)
    \end{tabularx}
    
    \noindent\begin{tabularx}{\linewidth}{@{}p{0.25cm}YYY@{}}
{\adjustbox{angle=90, raise=1.3cm-0.5\height}{DSM geometry}} &
\includegraphics[width=2.6cm, trim=10mm 10mm 10mm 10mm, clip]{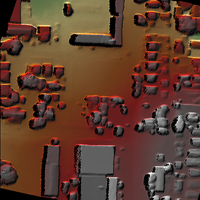} &
\includegraphics[width=2.6cm, trim=10mm 10mm 10mm 10mm, clip]{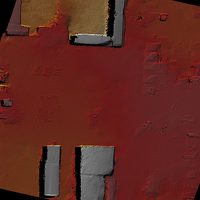} &
\includegraphics[width=2.6cm, trim=10mm 10mm 10mm 10mm, clip]{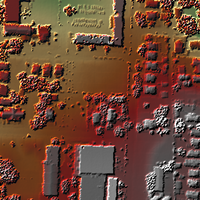} \\
& Ours (\maeOurs\,m) & Zero-shot (\maeZero\,m) & LiDAR GT
\end{tabularx}
}
    \caption{%
    Output geometry for a winter-autumn image pair from %
    Omaha (OMA\_331 test scene). 
    Our method recovers accurate geometry despite the diachronic nature of the pair, exhibiting strong appearance changes, which cause existing zero-shot methods to fail.
    Missing values due to perspective shown in black. Mean altitude error in parentheses; lower is better.}
    \label{fig:teaser}
\end{figure}

\section{Introduction}\label{INTRODUCTION}

Stereoscopic reconstruction models are becoming increasingly powerful, with the latest advances in the state of the art driven by the integration of monocular priors~\cite{cheng2025monster,bartolomei2025stereo,wen2025foundationstereo}. In the remote sensing community, progress often builds on these general advances, with models adapted %
to the particular requirements of satellite imagery.
After fine-tuning on the target domain, these models consistently achieve superior performance, surpassing classical techniques.
Deep stereo matching architectures now outperform long-standing semi-global matching algorithms \cite{hirschmuller2007stereo,facciolo2015mgm,mari2022disparity}, commonly used in satellite stereoscopic pipelines, such as S2P~\cite{amadei2025s2p,de2014automatic} and ASP~\cite{beyer2018ames}.

Despite this progress, significant challenges remain. As highlighted in~\cite{tosi2025survey}, one persistent difficulty arises in scenarios with challenging weather conditions. In Earth observation systems, diachronic stereoscopic pairs, i.e., images of the same geographical area captured at different dates with different viewing angles, are common. In contrast, true stereo acquisitions from the same date are more costly and less frequent. Consequently, addressing stereo reconstruction from diachronic images is of practical importance. As noted in \cite{facciolo2017automatic,mari2022disparity}, performance deteriorates as the temporal gap between image pairs increases. This degradation becomes particularly severe under substantial seasonal differences, such as snow versus no-snow conditions, where both traditional and learning-based methods fail dramatically. Moreover, as discussed in \cite{amadei2025s2p}, pipelines originally designed for near-simultaneous imagery struggle with the specific challenges of multi-temporal stereo, including variations in lighting, shadows, weather conditions, and moving objects, all of which adversely affect the accuracy of geometric reconstruction.

In this work, we present the first model capable of reliably performing diachronic satellite stereo matching. To enable this, we fine-tune a pretrained MonSter model~\cite{cheng2025monster} on a stereoscopic dataset derived from Track 3 of the \textit{DFC2019} remote sensing challenge~\cite{dfc1,dfc2}, specifically curated to include both synchronic and diachronic pairs selected according to temporal and geometric criteria. Each pair was rectified to ensure consistent disparity direction and alignment with reference Digital Surface Models (DSMs) derived from the LiDAR ground truth. We chose MonSter as it leverages monocular depth estimates, which are largely invariant to appearance changes unrelated to geometric structure.

Our approach achieves state-of-the-art performance in terms of the mean absolute error (MAE) between the reference and reconstructed DSMs, derived from the estimated disparities. We demonstrate this through an extensive evaluation, using pairs from the \textit{DFC2019 Track~3} data, covering areas in Jacksonville and Omaha (USA), and pairs from the \textit{IARPA~2016} dataset~\cite{iarpa}, covering areas near Buenos Aires (Argentina). All models are trained exclusively on a subset of \textit{DFC2019~Track~3}, and tested on previously unseen sites, such as the Buenos Aires areas. The \textit{IARPA~2016} data and a subset of \textit{DFC2019} Jacksonville scenes (specifically JAX\_004, JAX\_068, JAX\_214, and JAX\_260) have become benchmark sites for satellite multi-view reconstruction, used to evaluate recent methods such as EO-NeRF~\cite{mari2023multi} and EOGS~\cite{aira2025gaussian}. In our evaluation, Jacksonville scenes serve as a reference for synchronic stereo, while Omaha and Buenos Aires sites provide more challenging diachronic conditions, characterized by large temporal gaps and substantial appearance variations.

Across all these datasets, we observe a clear performance hierarchy: zero-shot state-of-the-art models, such as MonSter~\cite{cheng2025monster}, FoundationStereo~\cite{wen2025foundationstereo}, and StereoAnywhere~\cite{bartolomei2025stereo}, underperform relative to models fine-tuned on domain-specific data. Fine-tuning MonSter only on synchronic pairs leads to significant improvements, while fine-tuning on combined synchronic–diachronic pairs yields the best overall results. Our fine-tuned model also outperforms both the classical s2p-hd pipeline \cite{amadei2025s2p} and previous learning-based methods such as RAFT-Stereo \cite{lipson2021raft}, even when fine-tuned on our dataset. Furthermore, in aerial imagery benchmarks such as Enschede and EuroSDR-Vaihingen~\cite{wu2024evaluation}, our model performs on par with or surpasses competing methods.

In summary, this work introduces two key contributions:
\begin{itemize}
\setlength{\topsep}{0.1cm}
\setlength{\itemsep}{0.1cm}
\setlength{\parsep}{0pt}
\setlength{\partopsep}{0pt}
    \item We present the first model capable of reliably performing diachronic stereo matching, enabling 3D reconstruction from pairs of satellite images acquired at distant dates and under strong appearance changes. We show that both the inclusion of diachronic data during training and the use of monocular depth priors are key for achieving these results.
    \item We release\footnote{\url{https://centreborelli.github.io/diachronic-stereo}} a curated dataset for stereo matching in satellite imagery, including ground-truth disparities and DSMs, organized into synchronic and diachronic pairs. We also propose a simple heuristic based on metadata and image appearance to label pairs as diachronic or synchronic.
\end{itemize}

\section{Related Work}\label{RELATED WORK}

\subsection{Stereoscopic Reconstruction}
We refer the reader to the comprehensive survey by~\cite{tosi2025survey} for an in-depth review of the most recent advances in stereo matching. The past five years have seen remarkable progress, driven by key architectural designs and novel paradigms for addressing open challenges such as domain generalization, accuracy, over-smoothing of predictions, and efficiency.

Starting from RAFT-Stereo~\cite{lipson2021raft}, iterative optimization-based architectures have achieved impressive results in stereoscopic reconstruction. IGEV~\cite{xu2023iterative} further advanced RAFT's iterative cost-volume refinement paradigm, reaching state-of-the-art performance and later serving as the backbone of MonSter~\cite{cheng2025monster}.

Beyond architectural innovations, recent years have also been characterized by the extensive use of synthetic data. Synthetic datasets, e.g. \cite{mayer2016large}, have played a crucial role in the pre-training of stereo networks~\cite{tosi2025survey}. Although zero-shot adaptation to real imagery was initially challenging, recent works trained exclusively on synthetic data have demonstrated remarkable performance and strong generalization to real-world settings~\cite{lipson2021raft,xu2023iterative,tosi2023nerf,wen2025foundationstereo,cheng2025monster,bartolomei2025stereo}. It is also worth noting that the final boost in accuracy is commonly achieved when fine-tuning on the real target domain becomes feasible.

Recently, a new wave of stereo matching models has emerged that leverage monocular depth priors to improve performance in ill-posed scenarios such as occlusions and textureless regions~\cite{cheng2025monster,wen2025foundationstereo,bartolomei2025stereo}. Of particular relevance to this work is MonSter~\cite{cheng2025monster}, which we adopt as our base model, as it is the only one that has released the complete training code.

As noted by~\cite{tosi2025survey}, despite the remarkable progress achieved in recent years, several challenges remain open. Among them are handling very high-resolution imagery, coping with challenging weather conditions, addressing ill-posed scenes, and recovering fine structural details.

\subsection{Stereovision for Satellite Imagery}

Recent advances in 3D reconstruction from multiple images have been driven by Neural Radiance Fields (NeRF)~\cite{mildenhall2021nerf} and 3D Gaussian Splatting (3DGS)~\cite{kerbl20233d}.
In the context of satellite imagery, this trend is no exception. Recent works have adapted these approaches to the satellite domain, including EO-NeRF~\cite{mari2023multi} and EOGS~\cite{aira2025gaussian}, based on NeRF and 3DGS, respectively. These methods achieve remarkably strong results, both in appearance modeling and 3D reconstruction, but also have limitations. While they can handle complex illumination changes, they fail to capture severe seasonal variations such as snow versus no-snow conditions. Moreover, their performance relies on the availability of multiple views, typically five or more, and degrades significantly when only a few images are available~\cite{zhang2023spsnerf,masquil2025s}.

On the other hand, when only two images are available and captured within a short time span, scalable stereo pipelines based on semi-global matching \cite{hirschmuller2007stereo} perform very well~\cite{amadei2025s2p,de2014automatic}. These approaches can be extended to multi-date configurations by pairing temporally close images and fusing the reconstructed DSMs to form a consistent 3D model~\cite{facciolo2017automatic,gomez2023improving}. Such pairwise fusion often yields better results than traditional multi-view methods~\cite{facciolo2017automatic}.

Recent studies, such as~\cite{wu2024evaluation}, have analyzed state-of-the-art learning-based stereo matching methods in the context of aerial imagery. They found that domain adaptation plays a crucial role, with substantial gains observed after fine-tuning models on the target domain. Similarly,~\cite{mari2022disparity} demonstrated that deep learning methods generally outperform classical matching-based algorithms in satellite stereo reconstruction under ideal conditions, and that pretrained aerial models can adapt well to satellite data. However, these models require careful input preprocessing to match training conditions and often produce incomplete reconstructions in complex or unusual scenarios (e.g., very distant acquisition dates or narrow baselines).

Existing datasets for stereo matching in satellite imagery have been reviewed in~\cite{patil2023stellar}, which also introduces a new large-scale benchmark. They highlight several limitations of previous datasets, including limited geographic coverage, lack of diversity, and most critically, the presence of \textit{bipolar} disparities, i.e., disparities with mixed signs, which are not compatible with the most recent approaches. Although their dataset effectively addresses many of these issues, it remains unsuitable for our purposes. In particular, their rectification process results in extremely wide disparity ranges, which must be constrained for deep stereo models to operate effectively and to enable manageable crop sizes, as their distributed image tiles are approximately $5000 \times 5000$ pixels. Moreover, the image pairs are provided without the corresponding camera models or rectification homographies, making it impossible to evaluate DSM reconstructions and restricting assessment to disparity errors alone.

\section{Fine-Tuning MonSter for Diachronic Stereo Matching}\label{METHOD}

We describe the methodology used to fine-tune MonSter for reliable diachronic stereo matching in satellite imagery. 
Sections~\ref{sec:data_curation}-\ref{sec:finetuning} cover the main stages of the training process: (i) dataset curation and RPC-based rectification of image pairs to enforce unipolar positive horizontal disparities and derive disparity supervision from DSMs; and (ii) fine-tuning a state-of-the-art stereo matching network (MonSter) on a balanced mix of synchronic and diachronic pairs. The predicted disparities are triangulated using the RPC models and projected to produce a DSM (Section~\ref{sec:dsm_reconstruction}). The reconstructed DSMs are used for evaluation.

\subsection{Dataset Curation}
\label{sec:data_curation}

We use the WorldView-3 RGB crops from Track~3 of the \textit{DFC2019} challenge~\cite{dfc1,dfc2} and their corresponding ground-truth DSMs built from LiDAR to derive disparities. The dataset comprises 110 areas of interest (AOIs), consisting of 54 in Jacksonville (JAX) and 56 in Omaha (OMA). On average, each JAX scene comprises approximately 20 images, while each OMA site features around 32.

In this work, we label image pairs according to the temporal gap between the two images and their visual similarity, measured by the number of SIFT~\cite{lowe2004sift} feature matches. 
This SIFT-based criterion is effective in practice, as these features are highly sensitive to appearance variations such as those caused by seasonal changes~\cite{mari2019bundle}. 
Thus, we define 
\begin{itemize}
\setlength{\topsep}{0.05cm}
\setlength{\itemsep}{0.05cm}
\setlength{\parsep}{0pt}
\setlength{\partopsep}{0pt}
    \item \emph{\bf Diachronic pairs} as pairs of satellite images of the same geographic area acquired more than 30~days apart (modulo~1~year), and exhibiting fewer than 40 SIFT matches between them.

   \item  \emph{\bf Synchronic pairs} as image pairs acquired within a 30-day interval and having at least 40 SIFT matches between them. We note that the 40-match threshold could be normalized with respect to the surface area to be image size agnostic.

\end{itemize}

To build a diverse dataset with an emphasis on diachronic conditions, we iterate over all AOIs and extract, for each one, at least 30~diachronic pairs in Omaha and 3~diachronic pairs in Jacksonville. Additionally, we randomly sample 5 synchronic pairs per site to balance the dataset. Diachronic effects are less frequent in Jacksonville, where vegetation and lighting remain more consistent throughout the year, resulting in more minor appearance differences even between acquisitions made several months apart. This \textit{diachronic+synchronic} dataset is the main training resource used in this work. Additionally, we build a \textit{synchronic-only} dataset with approximately 15 synchronic pairs per AOI, which is used for ablation studies.

For each selected pair, we rectify the images and compute a ground-truth disparity map using the reference DSM and RPC camera models, as detailed in Section~\ref{sec:rectification}. In total, we obtain 2{,}246 pairs in the diachronic+synchronic dataset and 1{,}567 pairs in the synchronic-only dataset.

\subsection{Rectification of Multi-Date Satellite Imagery}
\label{sec:rectification}

{
This section describes the procedure used to rectify multi-date satellite image pairs into a geometry compatible with stereo networks (Algorithm~\ref{algo:rectification_v2}) and the procedure for computing ground-truth disparities from DSMs for supervised training (Algorithm~\ref{algo:gt_disp}).

Deep stereo architectures, including MonSter, require disparities to be \emph{unipolar}-that is, pixels in the right image should appear shifted to the left relative to those in the left image-- and to \emph{increase} with the underlying altitude, so that higher elevations correspond to larger disparity values. Enforcing these geometric constraints is nontrivial, as it requires identifying at least a few reliable image matches, \emph{correspondences} across diachronic pairs, where substantial appearance changes due to season, illumination, or vegetation often make classical feature matching difficult.

To address this, we design a rectification strategy (Algorithm~\ref{algo:rectification_v2}) that transforms arbitrary multi-date pairs into rectified stereo pairs compatible with MonSter’s architecture and training distribution. The algorithm progressively refines a pair of rectifying homographies. It is initialized using RPC-based virtual correspondences following the approach of \cite{de2014automatic}, then refined by fitting a horizontal shear transformation to reduce the disparity range. Because virtual correspondences alone are insufficient to guarantee \emph{unipolar} disparities, a small number of actual matches are required across the diachronic pair. We employ DISK~\cite{tyszkiewicz2020disk} for keypoint detection and LightGlue~\cite{lindenberger2023lightglue} for feature matching with conservative matching settings, as this combination provides greater robustness to seasonal and illumination variations than traditional descriptors such as SIFT (Figure \ref{fig:sift_vs_lg}). Similar to~\cite{mari2022disparity}, these sparse matches are used to estimate a global horizontal translation (and, if necessary, a polarity swap) that enforces unipolar and altitude-increasing disparities. %
The resulting rectifying homographies are further refined in order to minimize any remaining vertical alignment error, yielding a final configuration that satisfies the geometric assumptions of modern deep stereo architectures even for strongly diachronic imagery.

For completeness, Algorithm~\ref{algo:gt_disp} outlines how we derive the reference disparities from the ground-truth DSM. This step, required for fine-tuning and benchmarking, reprojects each pixel in the left rectified view through the DSM and camera models to compute the corresponding horizontal displacement in the right view, producing accurate disparity supervision in rectified coordinates.

\begin{algorithm}[ht]
\caption{Diachronic rectification}
\label{algo:rectification_v2}
\begin{algorithmic}[1]
\Require Left image $I_L$ with its camera model $\mathrm{RPC}_L$, right image $I_R$ with its camera model $\mathrm{RPC}_R$, average altitude estimate $z_{\mathrm{avg}}$
\Ensure Rectified images $\hat{I}_L$ and $\hat{I}_R$, rectifying homographies $H_L$ and $H_R$
\LeftComment{Run \cite{de2014automatic} Algorithm 1.}
\State $(H_L,H_R)\!\gets\!\mathrm{RPC\_Rectification}(\mathrm{RPC}_L,\mathrm{RPC}_R)$
\LeftComment{Reduce disparity range with tilt, shear, and translation using RPC matches at $z_{avg}$}
\State $H_R\!\gets\!\mathrm{ReduceDispRange}(H_L,H_R,\mathrm{RPC}_L,\mathrm{RPC}_R,z_{\mathrm{avg}})$
\If{$\mathrm{DispDecreasesWithAltitude}(H_R,H_L, RPC_R, RPC_L)$}
    \LeftComment{Reverse the role of left and right images}
    \State $I_L,I_R\!\gets\!I_R,I_L$
    \State $H_L,L_R\!\gets\!H_R,H_L$
\EndIf
\State $(\hat{I}_L,\hat{I}_R)\!\gets\!\mathrm{Warp}(I_L,H_L),\mathrm{Warp}(I_R,H_R)$
\LeftComment{Extract robust keypoint matches}
\State $M\!\gets\!\mathrm{LightGlueMatching}(\hat{I}_L,\hat{I}_R)$
\LeftComment{Enforce positive unipolar disparities}
\State $t\!\gets\!\text{min}_{(u_L,v_L)\leftrightarrow(u_R,v_R)\in M}(u_L-u_R)$
\LeftComment{Correct residual vertical alignment errors}
\State $s\!\gets\!\text{median}_{(u_L,v_L)\leftrightarrow(u_R,v_R)\in M}(v_L-v_R)$
\State $H_R\!\gets\!\begin{bmatrix}1&0&t\\0&1&s\\0&0&1\end{bmatrix}H_R$
\State $(\hat{I}_L,\hat{I}_R)\!\gets\!\mathrm{Warp}(I_L,H_L),\mathrm{Warp}(I_R,H_R)$
\State \Return $\hat{I}_L, \hat{I}_R, H_L, H_R$
\end{algorithmic}
\end{algorithm}

\begin{figure}[h]
\centering
\setlength{\tabcolsep}{1pt}
\renewcommand{\arraystretch}{0.0}
\begin{tabular}{@{}cc@{}}
\adjustbox{angle=90, raise=0.225\columnwidth-0.5\height}{\scriptsize SIFT} &
\includegraphics[width=0.9\columnwidth]{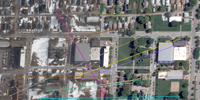}\\[2pt]
\adjustbox{angle=90, raise=0.225\columnwidth-0.5\height}{\scriptsize DISK + LightGlue} &
\includegraphics[width=0.9\columnwidth]{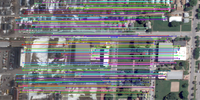} \\
\end{tabular}
\vspace{-0.5em}
\caption{Comparison of SIFT and DISK+LightGlue matches for the diachronic pair OMA\_331\_017 -- OMA\_331\_036.}
\label{fig:sift_vs_lg}
\end{figure}

\begin{algorithm}[ht]
\caption{Ground truth disparities computation}
\label{algo:gt_disp}
\begin{algorithmic}[1]
\Require Rectified left image $\hat{I}_L$, rectifying homographies $H_L$ and $H_R$, ground truth DSM, camera models $\mathrm{RPC}_L$ and $\mathrm{RPC}_R$
\Ensure Ground truth disparity $D$
\State $\text{DSM}_L \gets \text{ProjectDSM}(\text{DSM}, \text{RPC}_L)$ 
\For{each pixel $(u, v)$ in $\hat{I}_L$}
    \State $(x, y) \gets H_L^{-1}(u, v)$
    \State $z \gets \text{DSM}_L(x, y)$
    \State $(u_R, v_R) \gets H_R \circ \text{RPC}_R \circ \text{RPC}_L^{-1} (x, y, z)$
    \State $D(u, v) \gets u - u_R$
\EndFor

\State \Return $D$
\end{algorithmic}
\end{algorithm}

}

\subsection{Stereo Architecture}\label{STEREOMATCHING}
Our approach builds upon MonSter \cite{cheng2025monster}, a recent state-of-the-art stereo model that couples monocular priors with stereo matching. MonSter makes extensive use of the monocular depth model Depth Anything V2 \cite{yang2024depth}, employing it both as a backbone for feature extraction and as a depth estimator. The architecture consists of two main branches and a mutual refinement module. The stereo branch extracts features from the left and right images and follows the IGEV framework \cite{xu2023iterative} to produce an initial disparity estimate. The monocular branch predicts an initial relative depth map, which is converted into a metric monocular disparity using the initial stereo prediction. Finally, both the metric monocular depth and stereo disparities are mutually refined through an iterative refinement stage to obtain the final disparity estimate.

A key motivation for using monocular priors in diachronic stereo is their invariance to image appearance changes unrelated to geometric structure. Seasonal or illumination variations, such as those between winter and summer conditions or different shadows, can significantly alter pixel intensities corresponding to the same underlying geometry. While monocular depth models cannot produce metrically consistent values across images, they preserve the geometric structure of the scene. As a result, the features extracted by such models remain stable under varying conditions, providing strong and complementary cues for reliable correspondence estimation in diachronic matching. In our experiments, we observed that the monocular predictions of the same area across different seasons exhibit similar geometric structure despite large radiometric differences between the input images (cf. Figure~\ref{fig:monocular_consistency}).

\newlength{\panelSize}
\setlength{\panelSize}{0.315\columnwidth} %
\begin{figure}[t]
    \centering

\setlength{\tabcolsep}{1pt}
\small
    \begin{tabular}{cccc}
         &  RGB & Monocular Depth & LiDAR Altitude \\
         \adjustbox{angle=90, raise=0.5\panelSize-0.5\height}{Winter} &
        \includegraphics[width=\panelSize, height=\panelSize]{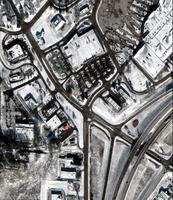} &
        \includegraphics[width=\panelSize, height=\panelSize]{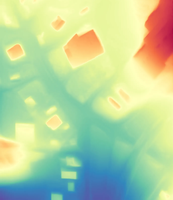}& 
         \includegraphics[width=\panelSize, height=\panelSize]{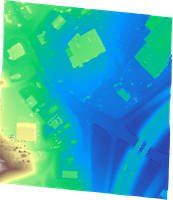}  \\
         \adjustbox{angle=90, raise=0.5\panelSize-0.5\height}{Summer} &
       \includegraphics[width=\panelSize, height=\panelSize]{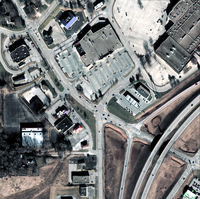}&
        \includegraphics[width=\panelSize, height=\panelSize]{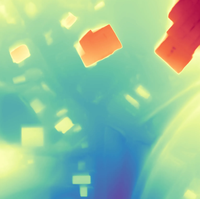}&
        \includegraphics[width=\panelSize, height=\panelSize]{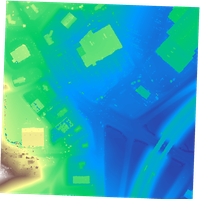}
    \end{tabular}
    
\vspace{-0.9em}
    \caption{Monocular depth consistency across seasons. Despite large radiometric differences between seasonal images from the Omaha OMA\_893 scene, the monocular Depth~Anything~V2 predictions exhibit coherent geometric structure.}
    \label{fig:monocular_consistency}
\end{figure}

\subsection{Fine-Tuning Details}
\label{sec:finetuning}

We fine-tune the MonSter model starting from its publicly available \textit{mix\_all} checkpoint, pretrained on a combination of the KITTI 2012 \cite{geiger2012we}, KITTI 2015 \cite{menze2015object}, Middlebury \cite{scharstein2014high}, ETH3D \cite{schops2017multi}, and SceneFlow \cite{mayer2016large} datasets. The model is optimized for 50,000 iterations using the AdamW optimizer \cite{loshchilov2017decoupled} with a learning rate of $5 \times 10^{-4}$ and a weight decay of $1 \times 10^{-5}$. Fine-tuning was performed on a single NVIDIA RTX 6000 GPU with a total training time of approximately 24 hours.

Input images are kept in the original dynamic range of \([0, 255]\) and randomly cropped to \(512 \times 512\) pixels for training. To prevent edge effects, a 32-pixel border is excluded from the loss computation. The fine-tuning configuration follows the default MonSter setup, including the
data augmentation strategies derived from RAFT-Stereo. Model performance is monitored using the absolute disparity error (end-point error) on the validation set, and the best checkpoint is selected based on this metric.

\subsection{DSM Reconstruction}
\label{sec:dsm_reconstruction}

Given a predicted disparity map, we want to reconstruct a DSM. First, we estimate an altitude value for each pixel in the rectified left view, using the iterative triangulation strategy described in Algorithm 2 of \cite{de2014automatic}. This algorithm searches for the height that best satisfies the epipolar constraint between the two RPC projections. We refer to this collection of all the estimated altitude values as an \textit{altitude image}. We then reproject the altitude image into a uniform ground-aligned grid using the left RPC model, obtaining a DSM.

\begin{table*}[t]
\centering
\setlength{\tabcolsep}{2pt}
\renewcommand{\arraystretch}{1.1}
\begin{tabular}{@{}l@{\hskip 0.3cm} l W{c}{2cm} W{c}{2cm} W{c}{2.5cm} W{c}{2.5cm}@{}}
\toprule
Method type & Model {\footnotesize{(checkpoint) \{fine-tuning\}}} & Jacksonville & Buenos Aires & Omaha Synchronic & Omaha Diachronic \\
\midrule
classical pipeline & s2p-hd & 2.04 $\pm$ 0.69 & 2.64 $\pm$ 0.78  & 1.33 $\pm$ 0.65 & 7.90 $\pm$ 3.68 \\ \midrule
 \multirow[c]{2}{*}{\begin{tabular}[c]{@{}l@{}} stereo networks\\{\footnotesize{(without monocular priors)}}\end{tabular}} & RAFT-Stereo {\footnotesize{(sceneflow)}}      & 6.01 $\pm$ 3.64 & 5.07 $\pm$ 1.36 & 1.44 $\pm$ 0.81 & 2.24 $\pm$ 0.99 \\ 
& RAFT-Stereo {\footnotesize{\{fine-tuned\}}}  & 1.73 $\pm$ 0.61 & 2.08 $\pm$ 0.32 & 0.89 $\pm$ 0.35 & 1.04 $\pm$ 0.44 \\
\midrule
\multirow[c]{5}{*}{\begin{tabular}[c]{@{}l@{}}stereo networks\\{\footnotesize{(with monocular priors)}}\end{tabular}} & FoundationStereo {\footnotesize{(23-51-11)}} & 2.33 $\pm$ 0.89 & 7.18 $\pm$ 4.41 & 1.42 $\pm$ 1.45 & 1.30 $\pm$ 0.95 \\
 & StereoAnywhere {\footnotesize{(sceneflow)}}   & 5.06 $\pm$ 2.04 & 3.53 $\pm$ 0.41  & 1.04 $\pm$ 0.44 & 1.38 $\pm$ 0.59 \\
 & MonSter {\footnotesize{(mix all)}}            & 2.15 $\pm$ 0.63 & 2.55 $\pm$ 0.45  & 0.92 $\pm$ 0.34 & 1.61 $\pm$ 0.63 \\
 & Ours {\footnotesize{\{\textit{synchronic-only}\}}}          & 1.29 $\pm$ 0.52 & 1.65 $\pm$ 0.10  & \textbf{0.76 $\pm$ 0.37} & 0.97 $\pm$ 0.44 \\
 & Ours {\footnotesize{\{\textit{diachronic+synchronic}\}}}    & \textbf{1.20 $\pm$ 0.52} & \textbf{1.55 $\pm$ 0.22} & 0.77 $\pm$ 0.33 & \textbf{0.84 $\pm$ 0.34}\\
\bottomrule
\end{tabular}
\caption{Evaluation on all satellite test sets for all methods: Altitude MAE in meters.}
\label{tab:satellite-results}
\end{table*}

\begin{table*}[h]
\centering
\small
\resizebox{\textwidth}{!}{
\begin{tabular}{l l c c c c c c c}
\toprule
AOI & Pair & \#SIFT &
FoundationStereo & StereoAnywhere & MonSter &
s2p-hd & Ours (Sync) & \textbf{Ours (Dia+Sync)} \\
\midrule
IARPA\_003 & 12JUN15-01SEP15 & 8 & 9.84 & 2.24 & 2.21 & 1.66 & 1.59 & \textbf{1.38} \\
JAX\_260 & 018-005 & 7 & 1.41 & 8.63 & 3.63 & 3.90 & 1.17 & \textbf{0.79} \\
OMA\_084 & 042-016 & 0 & 4.87 & 11.92 & 4.53 & 8.98 & 1.69 & \textbf{1.30} \\ 
OMA\_134 & 029-013 & 0 & \textbf{0.89}$^*$ & 2.82 & 3.58 & 16.97 & 2.39 & 1.33 \\
OMA\_230 & 013-031 & 0 & 0.68 & 8.51 & 1.21 & 12.33 & 0.46 & \textbf{0.34} \\
OMA\_247 & 009-023 & 7 & 1.10 & 1.29 & 3.08 & 12.21 & 0.86 & \textbf{0.66} \\
OMA\_331 & 017-036 & 7 & 1.75 & 3.10 & 2.68 & 4.14 & 1.22 & \textbf{0.88} \\
\bottomrule
\end{tabular}
}
\caption{Altitude MAE (m) for challenging diachronic test pairs exhibiting few SIFT matches. 
$^{*}$~FoundationStereo’s prediction for OMA\_134 is nearly flat, producing a deceptively small error despite a visually incorrect reconstruction, as shown in Figure \ref{fig:grid_oma}.}
\label{tab:grid_oma}
\end{table*}

\newcommand{\iarpaimg}[1]{\includegraphics[width=0.13\linewidth]{figures/4_experiments/big_grid/IARPA_003_12JUN15-01SEP15/#1}}
\newcommand{\jaximg}[1]{  \includegraphics[width=0.13\linewidth]{figures/4_experiments/big_grid/JAX_260_018-005/#1}}
\newcommand{\omaimg}[1]{  \includegraphics[width=0.13\linewidth]{figures/4_experiments/big_grid/OMA_247_009-023/#1}}

\begin{figure*}[t!]
\centering
\small
\setlength{\tabcolsep}{2pt}
\renewcommand{\arraystretch}{1}
\begin{tabular}{l@{\hskip 1pt}c@{\hskip 3pt}c@{\hskip 3pt}c@{\hskip 3pt}c@{\hskip 3pt}c@{\hskip 3pt}c@{\hskip 3pt}c}
 & {\scriptsize IARPA\_003} & {\scriptsize JAX\_260} & {\scriptsize OMA\_084} & {\scriptsize OMA\_134} & {\scriptsize OMA\_230} & {\scriptsize OMA\_247} & {\scriptsize OMA\_331} \\
\adjustbox{angle=90, raise=0.065\linewidth-0.5\height}{\scriptsize Left} &
\iarpaimg{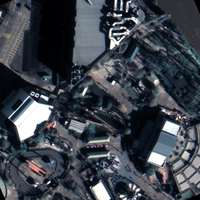} & 
\jaximg{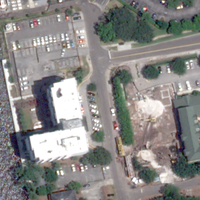} & 
\includegraphics[width=0.13\linewidth]{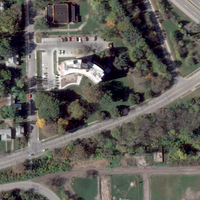} & 
\includegraphics[width=0.13\linewidth]{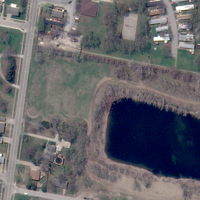} & 
\includegraphics[width=0.13\linewidth]{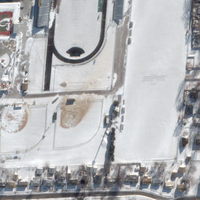} & 
\omaimg{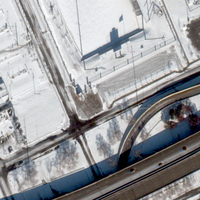} & 
\includegraphics[width=0.13\linewidth]{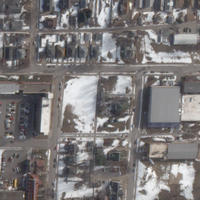} \\[-0.2em]
\adjustbox{angle=90, raise=0.065\linewidth-0.5\height}{\scriptsize Right} &
\iarpaimg{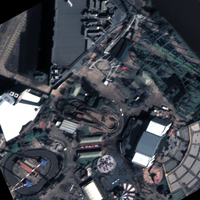} & 
\jaximg{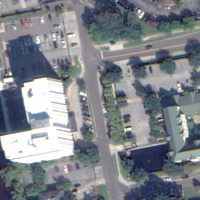} & 
\includegraphics[width=0.13\linewidth]{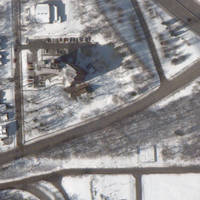} & 
\includegraphics[width=0.13\linewidth]{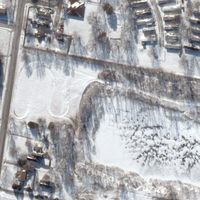} & 
\includegraphics[width=0.13\linewidth]{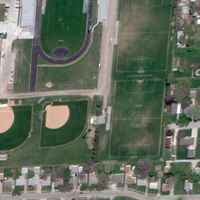} & 
\omaimg{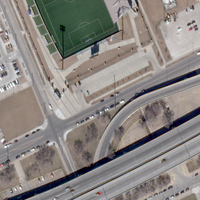} & 
\includegraphics[width=0.13\linewidth]{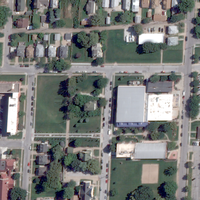} \\[-0.2em]
\adjustbox{angle=90, raise=0.065\linewidth-0.5\height}{\scriptsize FoundationStereo} & 
\iarpaimg{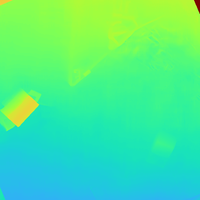} & 
\jaximg{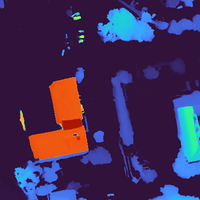} & 
\includegraphics[width=0.13\linewidth]{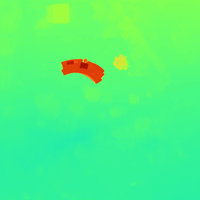} & 
\includegraphics[width=0.13\linewidth]{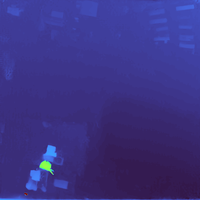} & 
\includegraphics[width=0.13\linewidth]{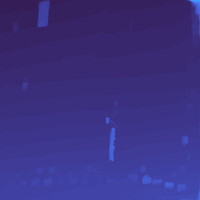} & 
\omaimg{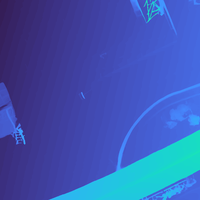} & 
\includegraphics[width=0.13\linewidth]{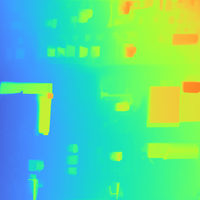} \\[-0.2em]
\adjustbox{angle=90, raise=0.065\linewidth-0.5\height}{\scriptsize StereoAnywhere} & 
\iarpaimg{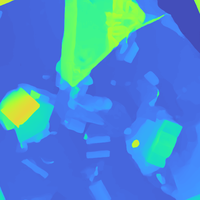} & 
\jaximg{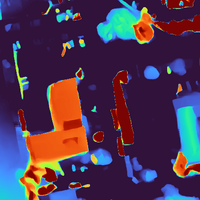} & 
\includegraphics[width=0.13\linewidth]{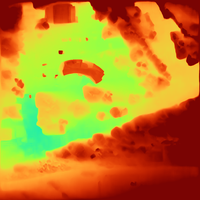} & 
\includegraphics[width=0.13\linewidth]{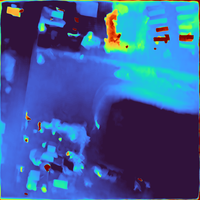} & 
\includegraphics[width=0.13\linewidth]{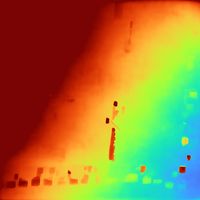} & 
\omaimg{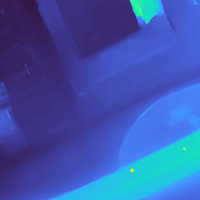} & 
\includegraphics[width=0.13\linewidth]{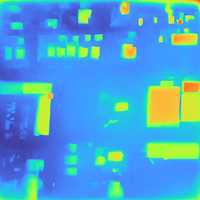} \\[-0.2em]
\adjustbox{angle=90, raise=0.065\linewidth-0.5\height}{\scriptsize MonSter} & 
\iarpaimg{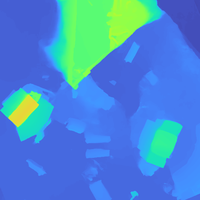} & 
\jaximg{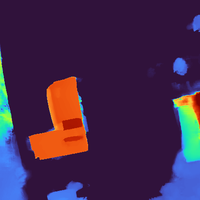} & 
\includegraphics[width=0.13\linewidth]{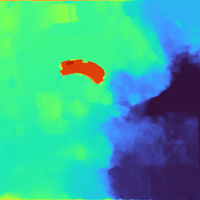} & 
\includegraphics[width=0.13\linewidth]{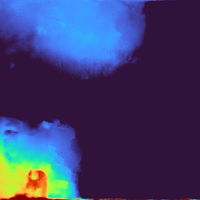} & 
\includegraphics[width=0.13\linewidth]{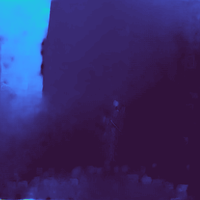} & 
\omaimg{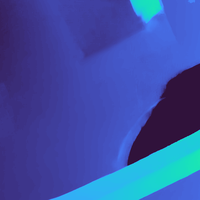} & 
\includegraphics[width=0.13\linewidth]{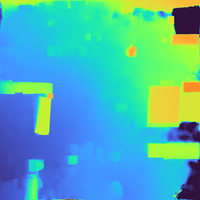} \\[-0.2em]
\adjustbox{angle=90, raise=0.065\linewidth-0.5\height}{\scriptsize Ours (Sync)} & 
\iarpaimg{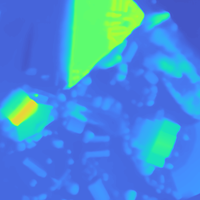} & 
\jaximg{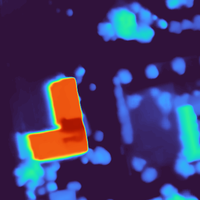} & 
\includegraphics[width=0.13\linewidth]{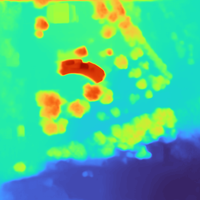} & 
\includegraphics[width=0.13\linewidth]{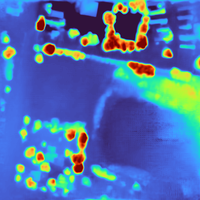} & 
\includegraphics[width=0.13\linewidth]{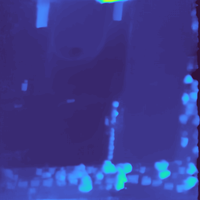} & 
\omaimg{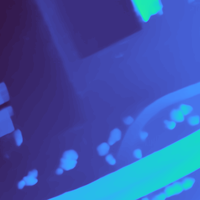} & 
\includegraphics[width=0.13\linewidth]{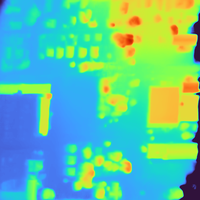} \\[-0.2em]
\adjustbox{angle=90, raise=0.065\linewidth-0.5\height}{\scriptsize Ours (Dia+Sync)} & 
\iarpaimg{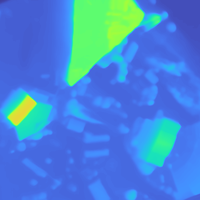} & 
\jaximg{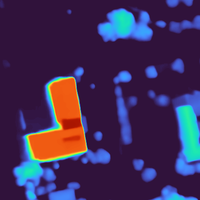} & 
\includegraphics[width=0.13\linewidth]{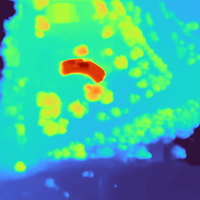} & 
\includegraphics[width=0.13\linewidth]{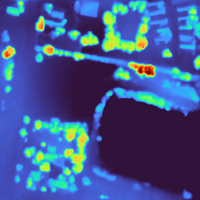} & 
\includegraphics[width=0.13\linewidth]{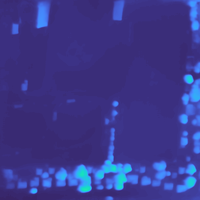} & 
\omaimg{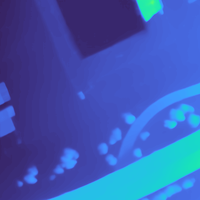} & 
\includegraphics[width=0.13\linewidth]{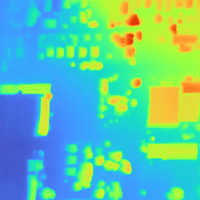} \\[-0.2em]
\adjustbox{angle=90, raise=0.065\linewidth-0.5\height}{\scriptsize GT} & 
\iarpaimg{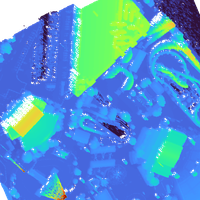} & 
\jaximg{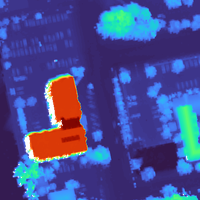} & 
\includegraphics[width=0.13\linewidth]{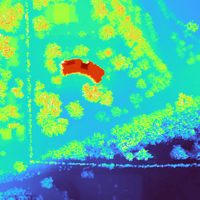} & 
\includegraphics[width=0.13\linewidth]{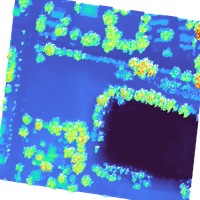} & 
\includegraphics[width=0.13\linewidth]{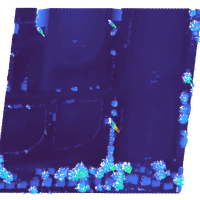} & 
\omaimg{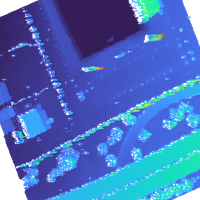} & 
\includegraphics[width=0.13\linewidth]{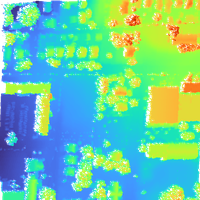} \\
\end{tabular}
\caption{Qualitative results of disparity predictions on a selection
of hard diachronic image pairs
from the test set,
listed in Table~\ref{tab:grid_oma}.}
\label{fig:grid_oma}

\end{figure*}
\begin{figure*}[!ht]
\centering
\small
\setlength{\tabcolsep}{2pt}
\renewcommand{\arraystretch}{1}
\begin{tabular}{l@{\hskip 1pt}c@{\hskip 3pt}c@{\hskip 3pt}c@{\hskip 3pt}c@{\hskip 3pt}c@{\hskip 3pt}c@{\hskip 3pt}c}
 & {\scriptsize IARPA\_003} & {\scriptsize JAX\_260} & {\scriptsize OMA\_084} & {\scriptsize OMA\_134} & {\scriptsize OMA\_230} & {\scriptsize OMA\_247} & {\scriptsize OMA\_331} \\
\adjustbox{angle=90, raise=0.065\linewidth-0.5\height}{\scriptsize s2p-hd} &
\includegraphics[width=0.13\linewidth]{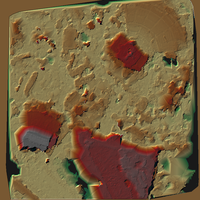} &
\includegraphics[width=0.13\linewidth]{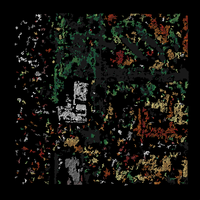} &
\includegraphics[width=0.13\linewidth]{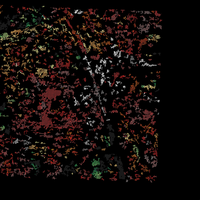} &
\includegraphics[width=0.13\linewidth]{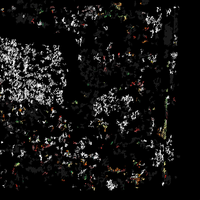} &
\includegraphics[width=0.13\linewidth]{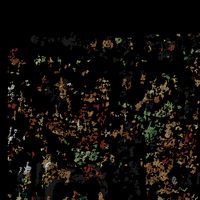} &
\includegraphics[width=0.13\linewidth]{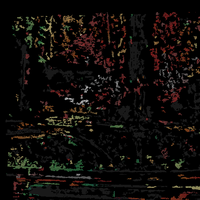} &
\includegraphics[width=0.13\linewidth]{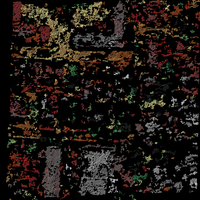} \\[-0.2em]
\adjustbox{angle=90, raise=0.065\linewidth-0.5\height}{\scriptsize MonSter} &
\includegraphics[width=0.13\linewidth]{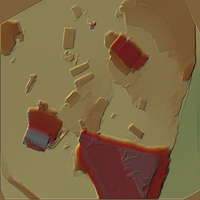} &
\includegraphics[width=0.13\linewidth]{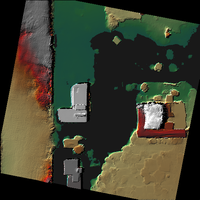} &
\includegraphics[width=0.13\linewidth]{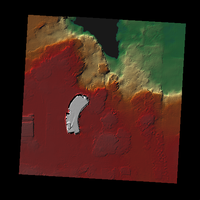} &
\includegraphics[width=0.13\linewidth]{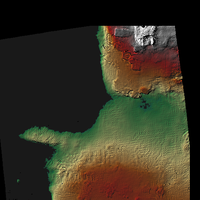} &
\includegraphics[width=0.13\linewidth]{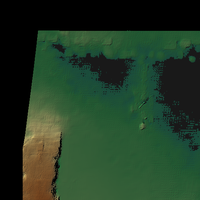} &
\includegraphics[width=0.13\linewidth]{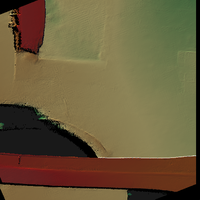} &
\includegraphics[width=0.13\linewidth]{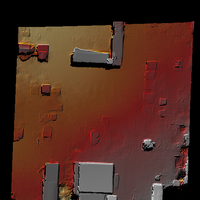} \\[-0.2em]
\adjustbox{angle=90, raise=0.065\linewidth-0.5\height}{\scriptsize Ours (Dia+Sync)} &
\includegraphics[width=0.13\linewidth]{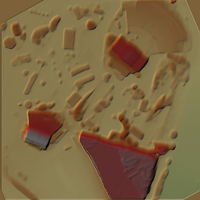} &
\includegraphics[width=0.13\linewidth]{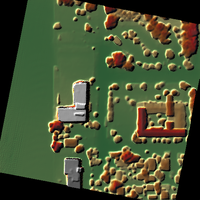} &
\includegraphics[width=0.13\linewidth]{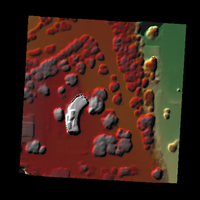} &
\includegraphics[width=0.13\linewidth]{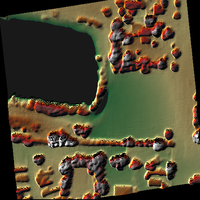} &
\includegraphics[width=0.13\linewidth]{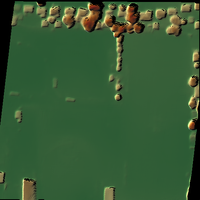} &
\includegraphics[width=0.13\linewidth]{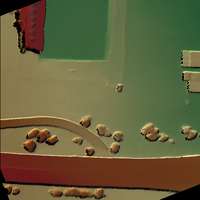} &
\includegraphics[width=0.13\linewidth]{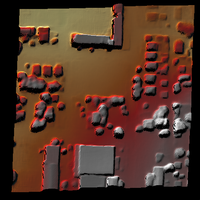} \\[-0.2em]
\adjustbox{angle=90, raise=0.065\linewidth-0.5\height}{\scriptsize GT} &
\includegraphics[width=0.13\linewidth]{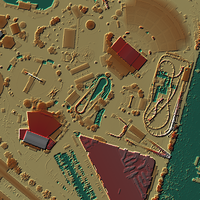} &
\includegraphics[width=0.13\linewidth]{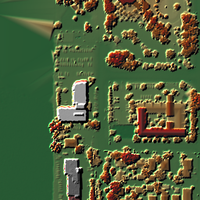} &
\includegraphics[width=0.13\linewidth]{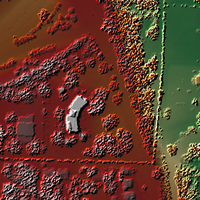} &
\includegraphics[width=0.13\linewidth]{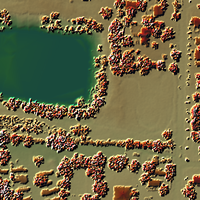} &
\includegraphics[width=0.13\linewidth]{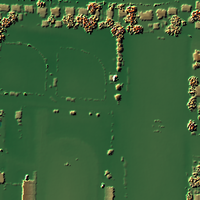} &
\includegraphics[width=0.13\linewidth]{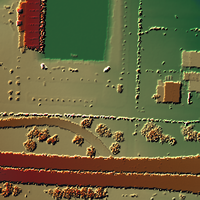} &
\includegraphics[width=0.13\linewidth]{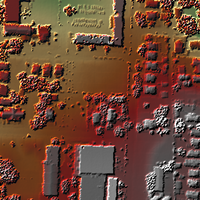} \\
\end{tabular}
\caption{Top to bottom: DSMs produced by the classical s2p-hd pipeline, zero-shot MonSter (mix all), our method, and the ground-truth DSM on a selection of hard diachronic image pairs
from the test set,
listed in Table~\ref{tab:grid_oma}. Missing values shown in black.}
\label{fig:grid_s2p_gt}
\end{figure*}

\section{Experiments}

We evaluate our approach on four test sets derived from DFC2019 Track 3 \cite{dfc1,dfc2} and the IARPA2016 dataset~\cite{iarpa}, commonly used in satellite multi-view reconstruction benchmarks:
\begin{itemize}
\setlength{\topsep}{0.05cm}
\setlength{\itemsep}{0.05cm}
\setlength{\parsep}{0pt}
\setlength{\partopsep}{0pt}
    \item Jacksonville (DFC2019): Four AOIs from Jacksonville, characterized by minimal seasonal or shadow changes between image pairs. This subset represents a synchronic test scenario.
    \item Buenos Aires (IARPA2016): Three AOIs from Buenos Aires with illumination and shadow differences between acquisitions. This represents a soft diachronic setting.
    \item Omaha Synchronic (DFC2019): Five AOIs from Omaha (OMA\_084, OMA\_134, OMA\_230, OMA\_247, OMA\_331) using only synchronic image pairs.
    \item Omaha Diachronic (DFC2019): The same five AOIs from Omaha, but using diachronic image pairs, showing strong seasonal contrasts between distant acquisitions, most notably snow versus no-snow conditions. This represents a strong diachronic scenario.
\end{itemize}

All models are trained on the Jacksonville and Omaha AOIs, excluding those reserved for testing. For each test AOI, we evaluate on twenty stereo pairs. For Jacksonville and Buenos Aires, pairs are randomly selected from all available combinations. For Omaha, we randomly pick 20 pairs for each test set according to the definitions from Section \ref{sec:data_curation}. All data and splits are shared to ensure reproducibility and to support future research.

Our primary metric is the pixelwise mean absolute error (MAE) between the reconstructed DSM and the LiDAR reference DSM. For each AOI, we evaluate it across multiple stereo pairs and calculate the median MAE as the AOI score. Since the difficulty varies significantly across regions, this aggregation produces a single, stable value that is robust to outliers and enables fair comparison across them. We then report the mean and standard deviation of these AOI scores as the score of each dataset.

Since vegetation height varies seasonally and may differ between image acquisitions and LiDAR capture dates, we also exclude vegetation regions. These ignored pixels are obtained from classification masks of the DSM provided in the original datasets. The LiDAR DSMs have a spatial resolution of 30–50 cm/pixel. To avoid boundary artifacts, a 32-pixel margin is cropped from all DSMs before error and loss computations, during training and testing.

\subsection{Diachronic Fine-Tuning of MonSter}

Table \ref{tab:satellite-results} reports the quantitative results on all test sets using the classical s2p-hd pipeline, different state-of-the-art stereo models (RAFT-Stereo~\cite{lipson2021raft}, FoundationStereo~\cite{wen2025foundationstereo}, StereoAnywhere~\cite{bartolomei2025stereo}, MonSter~\cite{cheng2025monster}), and our MonSter model fine-tuned with and without diachronic pairs (\textit{diachronic+synchronic} vs \textit{synchronic-only}, respectively). All third-party methods are executed with their default configurations. Across all test sets, our model consistently achieves the lowest MAE, demonstrating its robustness to diachronic conditions. Reduced variance indicates improved reliability and robustness, with no catastrophic errors. 

Figures \ref{fig:grid_oma} and \ref{fig:grid_s2p_gt} show qualitative results comparing our fine-tuned model against other zero-shot and fine-tuned baselines and s2p-hd, on a selection of test pairs. For each Omaha and Buenos Aires AOI, we deliberately select pairs with minimal SIFT feature matches, which correlate with severe appearance changes and thus represent the most challenging diachronic scenarios. The zero-shot stereo networks and s2p-hd fail under substantial seasonal variations, producing noisy (s2p-hd), incomplete (MonSter, StereoAnywhere) or over-smoothed surfaces (FoundationStereo) whereas our approach maintains consistent accuracy. The quantitative metrics for these pairs are provided in Table~\ref{tab:grid_oma}.

\paragraph{Impact of Monocular Priors}

To assess the contribution of monocular priors, we compare our method with a similarly fine-tuned RAFT-Stereo baseline using the \textit{diachronic+synchronic} training data. RAFT-Stereo~\cite{lipson2021raft} serves as the foundation for several recent architectures~\cite{cheng2025monster,bartolomei2025stereo}. However, unlike MonSter, it does not incorporate any monocular depth prior. Both models are fine-tuned using identical settings, data splits, and training duration to ensure a fair comparison. As shown in Table~\ref{tab:satellite-results}, our fine-tuned MonSter significantly outperforms the fine-tuned version of RAFT-Stereo. 

\paragraph{Impact of Diachronic Training Data}
We further analyze the role of diachronic training data by comparing
two variants of our model: one fine-tuned exclusively on the synchronic-only dataset and another fine-tuned on the combined diachronic+synchronic dataset. Results reported in Table~\ref{tab:satellite-results} indicate that incorporating diachronic pairs into the training set has a twofold effect. (i) It is non-harmful when evaluating on synchronic pairs, as demonstrated by the Jacksonville split and the Omaha synchronic split. (ii) It is essential when evaluating on diachronic pairs, as evidenced by the Omaha diachronic split.

\subsection{Generalization to Aerial Data}
In this section, we evaluate whether fine-tuning on diachronic satellite data improves generalization to a different domain: aerial imagery. We compare MonSter and our fine-tuned model on two datasets from the benchmark proposed by~\cite{wu2024evaluation}: Enschede and EuroSDR-Vaihingen.

For these datasets, we report errors in pixels for the disparity estimation, including both MAE and root mean square error (RMSE). The EuroSDR-Vaihingen dataset represents a relatively simple aerial stereo benchmark. Conversely, Enschede poses a more complex challenge, as noted by the benchmark authors~\cite{wu2024evaluation}, with all methods showing higher error levels.

As shown in Table~\ref{tab:aerial}, MonSter performance is comparable to our fine-tuned model in EuroSDR-Vaihingen. However, in the Enschede dataset, fine-tuning on diachronic satellite data reduces the domain gap, yielding improvements in MAE and RMSE. 

\begin{table}[!ht]
\centering
\resizebox{\columnwidth}{!}{%
\begin{tabular}{lcccc}
\toprule
\multirow{2}{*}{Model} & \multicolumn{2}{c}{EuroSDR-Vaihingen} & \multicolumn{2}{c}{Enschede} \\
\cmidrule(lr){2-3} \cmidrule(lr){4-5}
 & MAE (px) & RMSE (px) & MAE (px) & RMSE (px) \\
\midrule
MonSter (zero-shot) & \textbf{1.32} & 2.81 & 5.04 & 12.64 \\
\textbf{Ours (DFC fine-tuned)} & 1.57 & \textbf{2.74} & \textbf{3.93} & \textbf{9.35} \\
\bottomrule
\end{tabular}%
}
\caption{Evaluation on aerial datasets: MAE and RMSE in pixels.}
\label{tab:aerial}
\end{table}

\subsection{Limitations}
Despite the strong performance achieved by fine-tuning, our model still exhibits certain limitations inherent to learning-based methods. Once stereo matching becomes data-driven, the model inevitably learns to reproduce patterns present in the training data. Any systematic bias in the ground truth disparity maps is likely to appear at inference time.

\begin{figure}[t]
\centering
\setlength{\tabcolsep}{1pt}
\begin{tabular}{cccc}
\footnotesize Left & \footnotesize Right & \footnotesize With trees & \footnotesize Without trees \\ \\[-0.3cm]
\includegraphics[width=0.24\columnwidth]{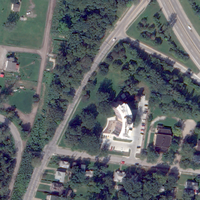} &
\includegraphics[width=0.24\columnwidth]{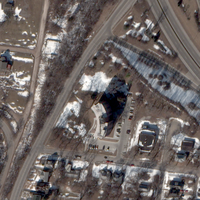} &
\includegraphics[width=0.24\columnwidth]{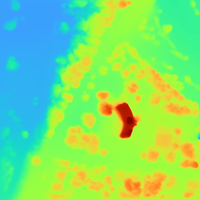} &
\includegraphics[width=0.24\columnwidth]{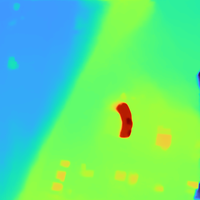} \\ \\[-0.55cm]
\includegraphics[width=0.24\columnwidth,angle=180]{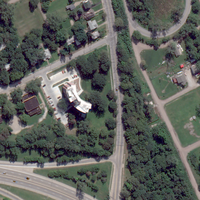} &
\includegraphics[width=0.24\columnwidth,angle=180]{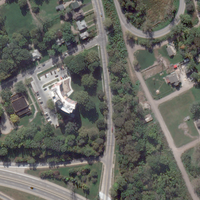} &
\includegraphics[width=0.24\columnwidth,angle=180]{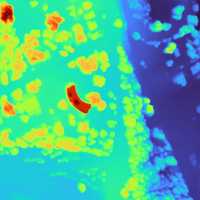} &
\includegraphics[width=0.24\columnwidth,angle=180]{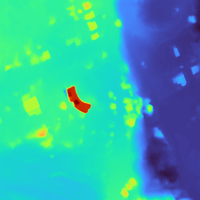} 

\end{tabular}
\caption{ %
Disparity predictions after fine-tuning on ground-truth disparity maps with vs without trees. Tree-inclusive fine-tuning is biased toward tree-inclusive predictions, while the opposite is biased toward vegetation-free predictions, regardless of whether trees are visible in one or both input images (top vs bottom). 
Tree-inclusive ground-truth disparities are shown in Figure \ref{fig:grid_oma}.
}
\label{fig:data_bias}
\end{figure}

\paragraph{Vegetation Bias} We trained an alternative version of our model using DSMs where vegetation pixels were replaced by the minimum height in their neighborhood, removing trees from the ground truth disparities. When comparing the two models, we observe that the model trained on the original, tree-inclusive data consistently predicts higher disparities corresponding to trees, even when they are not visible in both images. Conversely, the model trained without trees produces flat disparities in those regions, regardless of the trees appearing in both images. As shown in Figure~\ref{fig:data_bias}, this behavior is invariant to whether trees appear in one or both images.

\begin{figure}[t]
    \centering
\begin{tabular}{c@{\hspace{0.1cm}}c}
 \footnotesize Google Earth image (2013) & 
 \footnotesize LiDAR-derived GT disparity (2013) \\
\includegraphics[width=0.45\linewidth]{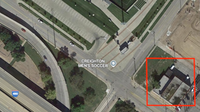} &
\includegraphics[width=0.45\linewidth]{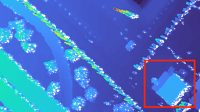} \\
\footnotesize Input image (2014, no building) & \footnotesize Disparity prediction \\
\includegraphics[width=0.45\linewidth]{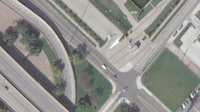} &
\includegraphics[width=0.45\linewidth]{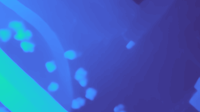}
\end{tabular}
    \caption{Temporal mismatch in OMA\_247 between images and  LiDAR-derived ground-truth disparity. The ground-truth (2013) includes a building absent from the input images (2014-2015). These inconsistencies add noise to both training and evaluation.}
    \label{fig:missing_building}
\end{figure}

\paragraph{LiDAR–Imagery Temporal Discordance} Using LiDAR-derived disparities as ground truth can introduce temporal noise that affects both training and evaluation. In Figure~\ref{fig:grid_oma}, the GT disparity for OMA\_247 contains a building that is absent in the images. This mismatch likely stems from LiDAR captured around 2013, before the structure was removed, while images were acquired between 2014 and 2015 \cite{dfc1,dfc2}. Figure~\ref{fig:missing_building} illustrates the discrepancy: the 2013 image (Google Earth) shows the building, while the 2014–2015 image does not. Although our model is generally robust to such inconsistencies during training, evaluation may be unfairly penalized, resulting in an increased reported error.

\newlength{\panelSizeTwo}
\setlength{\panelSizeTwo}{0.35\columnwidth}

\begin{figure}[t]
\centering
\begin{tabular}{@{}c@{\hspace{0.01\columnwidth}}c@{\hspace{0.01\columnwidth}}c@{}}
\includegraphics[width=0.31\columnwidth]{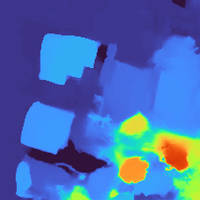} &
\includegraphics[width=0.31\columnwidth]{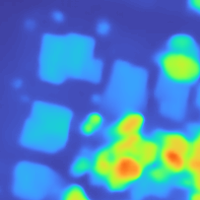} &
\includegraphics[width=0.31\columnwidth]{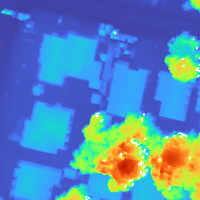} \\
\small \footnotesize Zero-shot MonSter & \footnotesize Fine-tuned MonSter & \footnotesize Ground truth \\
\footnotesize(MAE 1.07~m) & \footnotesize (MAE 0.72~m) & \footnotesize (LiDAR)
\end{tabular}
\caption{Comparison of disparity sharpness (detail from JAX\_004). Fine-tuning improves 
the accuracy of the predicted disparities but produces smoother, less sharp boundaries.}
\label{fig:sharpness}
\end{figure}

\paragraph{Reduced Sharpness} Another limitation caused by the training data is the reduced sharpness of the predicted disparities. As shown in Figure \ref{fig:sharpness}, zero-shot predictions are visually sharper, although they are less accurate overall. In contrast, our fine-tuned model produces more geometrically accurate results but with smoother edges. We attribute this loss of sharpness to the training disparities, which are derived from DSMs from LiDAR measurements. Although accurate, they may have limited details due to their spatial resolution. This contrasts with the original training data from MonSter, which consists of perfect disparities with sharp boundaries from purely synthetic data.

\section{Conclusion}

This work addressed the problem of diachronic stereo matching, i.e., recovering 3D geometry from multi-date satellite pairs affected by strong seasonal, illumination, and shadow changes. Our experiments reveal that deep stereo models, when carefully adapted, can solve this problem.

A key finding is that robustness to diachronic variations in this case does not necessitate architectural changes, but rather emerges from the data used during fine-tuning. We observe a clear performance hierarchy: zero-shot models perform worst, followed by models fine-tuned on synchronic satellite data only, while those trained on a combination of synchronic and diachronic pairs achieve the best results. Another key factor is the use of monocular priors, which help maintain geometric consistency when stereo cues might be weak due to appearance changes.

Nonetheless, our findings also expose persistent challenges common to learning-based stereo models. These models tend to reproduce the biases of their supervision data, e.g., predicting tree-like structures even when vegetation is not visible, or producing overly smooth results. While synthetic data could mitigate these effects by providing sharp and accurate supervision, capturing the full complexity of real-world settings remains an open challenge.

\section*{Acknowledgments}
This work was partially supported by Agencia Nacional de Investigacion e Innovacion (ANII, Uruguay) under the graduate scholarship POS\_NAC\_2023\_1\_177798.
This project was provided with computing HPC and storage resources by GENCI at CINES thanks to the grant AD011016256 on the supercomputer Adastra's MI250x partition.

{
	\begin{spacing}{1.17}
		\normalsize
		\bibliography{main} %
	\end{spacing}
}

\end{document}